\documentclass[sigconf]{acmart}

\AtBeginDocument{%
  \providecommand\BibTeX{{%
    \normalfont B\kern-0.5em{\scshape i\kern-0.25em b}\kern-0.8em\TeX}}}

\copyrightyear{2020}
\acmYear{2020}
\setcopyright{rightsretained} 

\acmConference[ICAIF '20]{ACM International Conference on AI in Finance}{October 15--16, 2020}{New York, NY, USA}
\acmBooktitle{ACM International Conference on AI in Finance (ICAIF '20),
October 15--16, 2020, New York, NY, USA}
\acmDOI{10.1145/3383455.3422565}
\acmISBN{978-1-4503-7584-9/20/10}

\usepackage{hyperref}       
\usepackage{url}            
\usepackage{booktabs}       
\usepackage{amsfonts}       
\usepackage{nicefrac}       
\usepackage{microtype}      
\usepackage{amsmath}
\usepackage{algorithm}
\usepackage{algorithmic}
\usepackage{graphicx}
\usepackage{subfigure}
\usepackage{amsmath}
\usepackage{amsthm}
\usepackage{blkarray}
\usepackage{commath}
\usepackage{adjustbox}
\usepackage{multirow}
\usepackage{float}
\usepackage{amsmath, nccmath}
\usepackage{lipsum}
\begin{document}

\title{Improved Predictive Deep Temporal Neural Networks \\ with Trend Filtering}

\author{Youngjin Park}
\email{youngjin@unist.ac.kr}
\affiliation{%
  \institution{\normalsize Ulsan National Institute of Science and Technology, UNIST}
  \streetaddress{50, UNIST-gil, Eonyang-eup, Ulju-gun}
  \city{Ulsan}
  \country{Republic of Korea}
  \postcode{44919}
}

\author{Deokjun Eom}
\email{deokjun.eom@kaist.ac.kr}
\affiliation{%
  \institution{\normalsize Korea Advanced Institute of Science and Technology, KAIST}
  \streetaddress{291, Daehak-ro, Yuseong-gu}
  \city{Daejeon}
  \country{Republic of Korea}
  \postcode{34141}
}
\author{Byoungki Seo}
\email{bkseo@unist.ac.kr}
\affiliation{%
  \institution{\normalsize Ulsan National Institute of Science and Technology, UNIST}
  \streetaddress{50, UNIST-gil, Eonyang-eup, Ulju-gun}
  \city{Ulsan}
  \country{Republic of Korea}
  \postcode{44919}
}
\author{Jaesik Choi}
\authornote{Jointly affiliated with INEEJI, Seongnam, Republic of Korea}
\email{jaesik.choi@kaist.ac.kr}
\affiliation{%
  \institution{\normalsize Korea Advanced Institute of Science and Technology, KAIST}
  \city{Daejeon}
  \country{Republic of Korea}
  \postcode{34141}
}

\begin{abstract}
    Forecasting with multivariate time series, which aims to predict future values given previous and current several univariate time series data, has been studied for decades, with one example being ARIMA. Because it is difficult to measure the extent to which noise is mixed with informative signals within rapidly fluctuating financial time series data, designing a good predictive model is not a simple task. Recently, many researchers have become interested in recurrent neural networks and attention-based neural networks, applying them in financial forecasting. There have been many attempts to utilize these methods for the capturing of long-term temporal dependencies and to  select more important features in multivariate time series data in order to make accurate predictions. In this paper, we propose a new prediction framework based on deep neural networks and a trend filtering, which converts noisy time series data into a piecewise linear fashion. We reveal that the predictive performance of deep temporal neural networks improves when the training data is temporally processed by a trend filtering. To verify the effect of our framework, three deep temporal neural networks, state of the art models for predictions in time series finance data, are used and compared with models that contain trend filtering as an input feature. Extensive experiments on real-world multivariate time series data show that the proposed method is effective and significantly better than existing baseline methods.

\end{abstract}


\maketitle

\section{Introduction}
Forecasting algorithms for multivariate time series have been widely applied in many areas, such as for, financial market prediction \cite{wu2013dynamic}, weather forecasting \cite{chakraborty2012fine}, groundwater level prediction \cite{sahoo2013groundwater}, and traffic predictions \cite{lai2018modeling}. There have also been a few methods for time series predictions such as ARIMA \cite{hillmer1982arima}, Gaussian Processes \cite{tong2019discovering, hwang2016automatic}. Despite of these efforts, it is still challenging to deal with data having complex and non-linear dependencies not only between time steps but also among the input features, which can change dynamically at each time step.

The recurrent neural network (RNN) \cite{rumelhart1986learning, werbos1990backpropagation, elman1991distributed}, a type of deep neural network designed for sequence modeling with shared trainable parameters, is actively used to solve prediction problems in sequential data to capture nonlinear relationships. Traditional RNNs, however, are associated with the vanishing or exploding gradient problems as well as some difficulty when used to detect long-term dependencies. Recently, long short-term memory (LSTM) \cite{hochreiter1997long} and the gated recurrent unit (GRU) \cite{cho2014learning} had some success in overcoming these limitations and haveachieved success in various areas, \textit{e.g.}, neural machine translation \cite{bahdanau2014neural}, and speech recognition \cite{graves2013speech}.  Attention mechanism based encoder-decoder structures \cite{cho2014learning, sutskever2014sequence} also helps to capture long-term dependencies by focusing on the selective parts of hidden states across all time steps.

Among the models based on LSTM or GRU with an attention mechanism, DA-RNN \cite{qin2017dual} is well designed with a two-stage attention architecture both to select appropriate input features and to capture the temporal sequence of the input data. A DA-RNN model uses exogenous information in which  the last time step T of the input features is included in the target time steps that the model is to predict. Unlike DA-RNN, DSANet \cite{huang2019dsanet} predicts future values without exogenous information and feeds each of the univariate time series data points independently into two parallel convolutional components before using self-attention modules \cite{uszkoreit2017transformer}. As exogenous information contains the future values of input features, all experiments are conducted without using exogenous information for all neural networks in this paper.

In time series finance data, considerable noise exists, and it is not easy to distinguish the noise from the important signals. Furthermore, structural break predictions are essential to estimate future values, but in noisy patterns of financial time series, recognizing structural breaks is challenging. This complicated structure of time series data disturbs proper predictions and drops the performances capabilities of deep neural networks. To alleviate these problems and to increase the performance for forecasting multivariate time series, we propose a method that adds a feature using L1 Trend Filtering (L1TF) \cite{kim2009ell_1} to the input values.

L1 trend filtering captures a linear pattern between two points of trend changes and uses a hyperparameter $\lambda$. Because $\lambda$, which can control the smoothness and residual, is optimally determined by numerous complicated experiments, the problem of the difficulty in distinguishing noise from informative signals can be solved to ensure that the model predicts appropriately. To confirm the effectiveness of this method, we use two cases, one that uses the trend filtering and the other that does not apply the filtering method, with stock index datasets with large numbers of driving series, \textit{e.g.}, the NASDAQ 100 Index, the EURO STOXX 60 Index, the Dow Jones Industrial Average, the FTSE 100 Index, and the TSX 60 Index. Extensive experiments on multivariate time series datasets demonstrate that the proposed method is effective and outperforms the baselines under paired t-test between models with and without the trend filtering.

\begin{figure*}[!ht]
\centering
\includegraphics[width=1\textwidth]{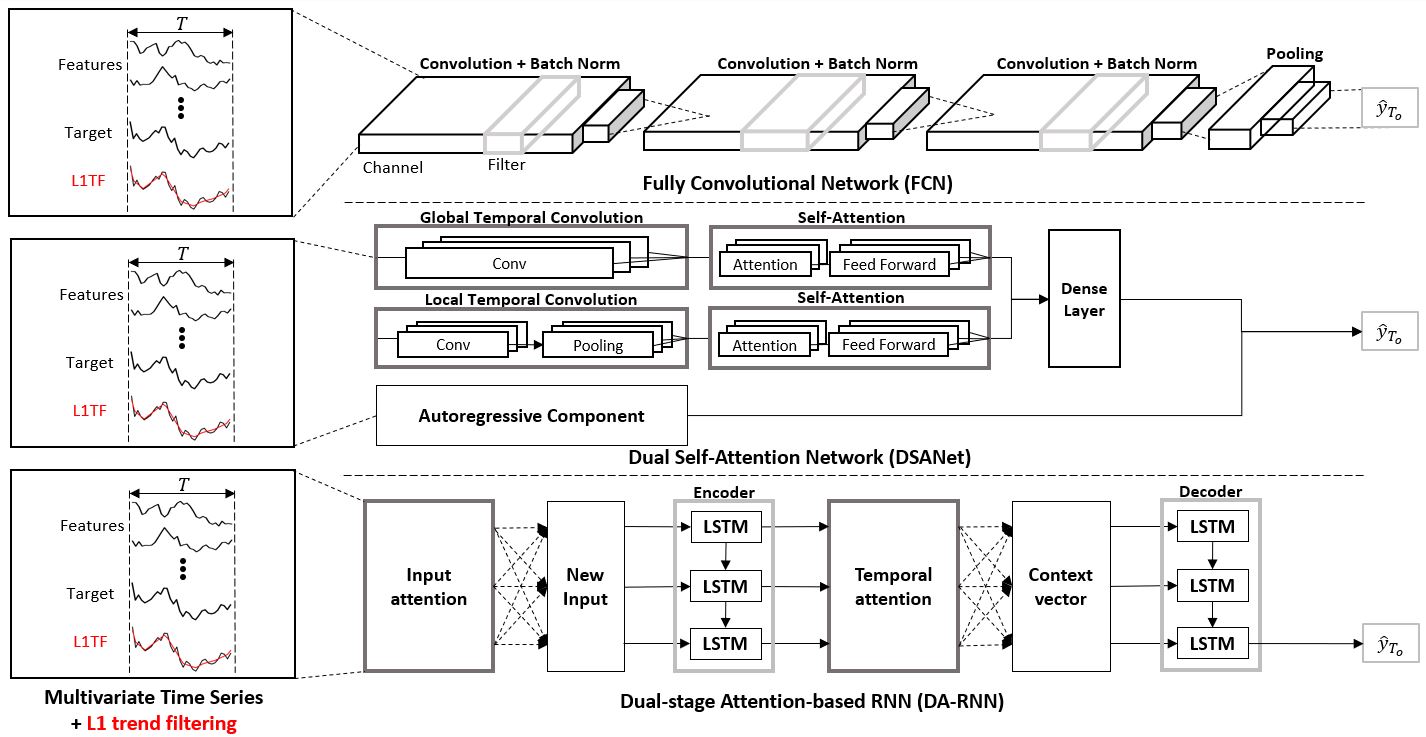}
\caption{Baseline model structures.}
\label{fig:model_structure}
\end{figure*}

\section{Background}
\subsection{Autoregressive Integrated Moving Average (ARIMA)}
By integrating an autoregressive (AR) with a moving average model (MA), the future value is assumed to be a linear function of several past observations and random errors. Therefore, the basic process that generates the time series has the following form:
\begin{equation}\label{eq:ARIMA}
\begin{aligned}
     y_t ={}& \theta_0 + \phi_1 y_{t-1}+\cdot \cdot \cdot +\phi_p y_{t-p}\\ 
      &  + \varepsilon_{t} - \theta_1 \varepsilon_{t-1} - \theta_2 \varepsilon_{t-2} - \cdot \cdot \cdot - \theta_q\varepsilon_{t-q},\quad d=0,
\end{aligned}
\end{equation}
where $y_t$ and $\varepsilon_{t}$ are the actual value and random error, respectively, at time $t$. $\phi_i$ $(i=1,2, ...,p)$ and $\theta_j$ $(j=0,1,2, ..., q)$ are the parameters of AR and MA estimated using an optimization procedure that minimizes the sum of square errors or some other appropriate loss function. $p$ and $q$ are integers that pertain to the order of the model. $d$ refers to differencing, which is used to calculate the difference between successive observations to indicate  the 
stationarity of the time series. One of the important aspects of the ARIMA model is that it determines the proper model hyperparameters $(p, d, q)$ and in this paper, we set d=0.

\subsection{Deep Neural Network}
DNN is a non-linear method that approximates the function $F$ that maps the input $\mathbf{X}$ to the output $\mathbf{y}$ as follows, 
\begin{equation}
\begin{split}
     \mathbf{y} &\approx F(X),
\end{split}
\end{equation}
 where $F$ represents the entire layer of DNN. We can then represent the equations of layers from the first layer to the output layer as follows, 

\begin{equation}\label{eq:Deep Neural Networks}
\begin{split}
    \mathbf{f}_1(\mathbf{X})&=g_1(W_1\mathbf{X}+\mathbf{b}_1), \\
    \mathbf{f}_2(\mathbf{X})&=g_2(W_{2}\mathbf{f}_{1}(\mathbf{X})+\mathbf{b}_2), \\
    \mathbf{f}_i(\mathbf{X})&=g_i(W_{i}\mathbf{f}_{i-1}(\mathbf{X})+\mathbf{b}_i), \\
    \mathbf{f}_{L-1}(\mathbf{X})&=g_{L-1}(W_{L-1}\mathbf{f}_{L-2}(\mathbf{X})+\mathbf{b}_{L-1}), \\
    \mathbf{y} &= \mathbf{f}_{L-1}(\mathbf{X})^TW_{L}+\mathbf{b}_{L},
\end{split}
\end{equation}
where $\mathbf{W_{i}}$ and $\mathbf{b_{i}}$ represent correspondingly the weight and bias for the $i$-th layer map, $\mathbf{f_{i}}$, while $g_{i}$ is an linear activation function for layer $i$. Typically, for each layer of a DNN, non-linear mapping is applied to the input with activation functions such as a logistic sigmoid, tanh, ReLU, and softmax.

\subsection{Fully Convolutional Network (FCN)}
A fully convolutional network (FCN) \cite{long2015fully, yi2017grouped} is popular in image semantic segmentation for taking input of an arbitrary size and producing correspondingly-sized output with efficient inference and learning. Each layer of data in a convolutional network has three dimensions, $h \times w \times d$, where $h$ and $w$ are spatial dimensions and $d$ denotes the number of channels. These basic components (convolution, batch normalization, pooling, and activation functions) are computed on local input regions.

In time series data, however, some basic components in a FCN must be changed to receive multivariate time series as input features. As it is not spatial, but temporal, the kernel size of one layer is $1 \times k$, where $k$ is the number of input time steps (window size) of one layer. $\mathbf{x}_t$ represents a vector of all input features at time t in a layer, and $\mathbf{y}_t$ is a vector for the following layer. This model is defined as follows,
\begin{equation}\label{eq:FCN}
\begin{split}
    \mathbf{y}_t = f_{ks}(\{\mathbf{x}_{st+\delta }\}_{0 \leq \delta \leq k}),
\end{split}
\end{equation}
where $k$ is the kernel size, $s$ is the stride, and $f_{ks}$ determines the layer type(i.e., a matrix multiplication for convolution, average pooling, a temporal max for max pooling). A FCN naturally operates on an input of any size and produces an output with corresponding temporal dimensions.

\subsection{Dual-stage Attention-Based Recurrent Neural Network (DA-RNN)}
A DA-RNN uses two attention mechanisms, an encoder with input attention to extract relevant driving series adaptively at each time step and a decoder with temporal attention to select relevant encoder hidden states across all time steps. Based on these mechanisms, the DA-RNN can capture relevant input features and the long-term temporal dependencies of a time series appropriately. This attention method can predict future target values effectively in multivariate time series finance data.

Given $\mathbf{n}$ driving series, $\mathbf{X}$ denotes input multivariate time series $i.e.$  $\mathbf{X} = (\mathbf{x}^1,\mathbf{x}^2, ..., \mathbf{x}^n)^{\top} = (\mathbf{x}_1, \mathbf{x}_2,..., \mathbf{x}_{T_i})\in\mathbb{R}^{n\times T_i}$, where $T_i$ is the length of the input window size and $n$ is the number of input features. Typically, the input index time series is represented as $(\mathbf{y}_1, \mathbf{y}_2, ..., \mathbf{y}_{T_i})$. The DA-RNN model aims to learn nonlinear mapping of the future values of the target series $\mathbf{y}_{T_o}$, where $T_o$ is defined as the length of the output time step size:
\begin{equation}\label{eq:DARNN}
\begin{aligned}
     \hat{\mathbf{y}}_{T_o} &\approx F(\mathbf{y}_1, ..., \mathbf{y}_{T_{i}}, \mathbf{x}_1, ..., \mathbf{x}_{T_{i}}), \quad T_o > T_i,
\end{aligned}
\end{equation}
where $F(\centerdot)$ is a nonlinear mapping function.

The sequence to sequence architecture is essentially an RNN that encodes the input sequence to a feature representation and decodes the representation to output the results. The DA-RNN model uses the LSTM structure as the encoder and decoder to capture long-term dependencies. Each LSTM unit has a memory cell with $\mathbf{s}_t$ at time $t$ and with hidden state $\mathbf{h}_t$ as the output. Access to the memory cell is controlled by three sigmoid gates: the forget gate $\mathbf{f}_t$, the input gate $\mathbf{i}_t$, and the output gate $\mathbf{o}_t$. The LSTM unit can be represented as follows,
\begin{equation}
\label{eq:LSTM}
\begin{gathered}
    \mathbf{f}_t = \sigma(W_f[\mathbf{h}_{t-1}; \mathbf{x}_t] + \mathbf{b}_f), \\
    \mathbf{i}_t = \sigma(W_i[\mathbf{h}_{t-1}; \mathbf{x}_t] + \mathbf{b}_i), \\
    \mathbf{o}_t = \sigma(W_o[\mathbf{h}_{t-1}; \mathbf{x}_t] + \mathbf{b}_o), \\
    \mathbf{s}_t = \mathbf{f}_t \odot \mathbf{s}_{t-1} + \mathbf{i}_t \odot \tanh{(W_s[\mathbf{h}_{t-1}; \mathbf{x}_t] + \mathbf{b}_s)}, \\
    \mathbf{h}_t = \mathbf{o}_t \odot \tanh{(\mathbf{s}_t)},
\end{gathered}
\end{equation}
where $[\mathbf{h}_{t-1}; \mathbf{x}_t]\in\mathbb{R}^{m+n}$ is a concatenation of the previous hidden state $\mathbf{h}_{t-1}\in\mathbb{R}^{m}$ and the current input $\mathbf{x}_t\in\mathbb{R}^{n}$. $W_f, W_i, W_o, W_s \in \mathbb{R}^{m\times(m+n)}$, and $\mathbf{b}_f, \mathbf{b}_i, \mathbf{b}_o, \mathbf{b}_s \in \mathbb{R}^m$ are trainable parameters.

Continuously, DA-RNN can construct an input attention mechanism via a deterministic attention model, $i.e.,$ a multilayer perceptron, by referring to the previous hidden state $\mathbf{h}_{t-1}$ and cell state $\mathbf{s}_{t-1}\in\mathbb{R}^{m}$ in the encoder LSTM unit with:
\begin{equation}\label{eq:DA-RNN/input_attention}
\begin{gathered}
    e_{t}^{k} = \mathbf{v}_{e}^{\top}\tanh{(W_e[\mathbf{h}_{t-1};\mathbf{s}_{t-1}] + U_{e}\mathbf{x}^k)},
\end{gathered}
\end{equation}
and
\begin{equation}\label{eq:DA-RNN/input_attention/softmax}
\begin{gathered}
    \alpha_t^k = \frac{exp(e_t^k)}{\sum_{i=1}^{n}exp(e_t^i)},
\end{gathered}
\end{equation}
where $T$ is the input time step, $\mathbf{x}^k\in\mathbb{R}^T, \mathbf{v}_e\in\mathbb{R}^T, W_e\in\mathbb{R}^{T\times2m}$, and $U_e\in\mathbb{R}^{T\times{T}}$ denotes the parameters to learn.

Temporal attention is used in the decoder, and the attention weight of each encoder hidden state at time t is calculated based upon the previous decoder hidden state $\mathbf{d}_{t-1}\in\mathbb{R}^p$ and cell state $\mathbf{s}'_{t-1}\in\mathbb{R}^p$ with:
\begin{equation}\label{eq:DA-RNN/temporal_attention}
\begin{gathered}
    l_{t}^{i} = \mathbf{v}_{d}^{\top}\tanh{(W_d[\mathbf{d}_{t-1};\mathbf{s}'_{t-1}] + U_{d}\mathbf{h}_i)}, \quad 1<i\leq T,
\end{gathered}
\end{equation}
and
\begin{equation}\label{eq:DA-RNN/temporal_attention/softmax}
\begin{gathered}
    \beta_t^i = \frac{exp(l_t^i)}{\sum_{j=1}^{T}exp(l_t^j)},
\end{gathered}
\end{equation}
where $[\mathbf{d}_{t-1};\mathbf{s}'_{t-1}]\in\mathbb{R}^{2p}$ is a concatenation of the previous hidden state and the cell state of the decoder LSTM unit, $\mathbf{v}_d\in\mathbb{R}^m, W_d\in\mathbb{R}^{m\times2p}$, and $U_d\in\mathbb{R}^{m\times{m}}$.
The attention mechanism of the DA-RNN can compute the context vector $\mathbf{c}_t$ as a weighted sum of all encoder hidden states ${\mathbf{h}_1, ..., \mathbf{h}_T}$,

\begin{equation}\label{eq:DA-RNN/context_vector}
\begin{gathered}
    \mathbf{c}_t = \sum_{i=1}^{T}\beta_t^i \mathbf{h}_i,
\end{gathered}
\end{equation}
where the context vector $c_t$ is distinct at each time step.
Therefore, $\hat{y}_{T_{o}}$ can be obtained with 
\begin{equation}\label{eq:DA-RNN/y_hat}
\begin{aligned}
    \hat{y}_{T_o} {} & = F(\mathbf{y}_1, ..., \mathbf{y}_T, \mathbf{x}_1, ..., \mathbf{x}_T) \\
    & = \mathbf{v}_{y}^{\top}(W_y\mathbf{d}_T + \mathbf{b}_w) + b_v,
\end{aligned}
\end{equation}
where the parameters $W_y\in\mathbb{R}^{p\times p}$ and $\mathbf{b}_w\in\mathbb{R}^p$ are mapping to the size of the decoder hidden states, and $\mathbf{v}_y\in\mathbb{R}^p$ and bias $b_v\in\mathbb{R}$ are the weights of a linear function, leading to the final prediction output. This part differs from the concepts in the DA-RNN paper. We did not use the context vector to create the prediction output due to the increased performance. 

\subsection{Dual Self-Attention Network (DSANet)}
DSANet can be applied to enable accurate and robust forecasting for multivariate time series. In this paper, DSANet also uses a problem statement identical to that in Eqn. \ref{eq:DARNN}.

DSANet utilizes two parallel convolutional components feeding each of the univariate time series independently, referred to as global temporal convolution and local temporal convolution respectively to capture complex mixtures of global and local temporal patterns.

Global temporal convolution uses a convolutional structure with multiple $T\times1$ filters to extract time-invariant patterns of all time steps for each driving time series. Each filter of the global temporal convolution module produces a vector with a size of $D\times1$, where the activation function is a ReLU function. Local temporal convolution uses the length of the filters as $l$, where $l<T$ to map local temporal relationships in each univariate time series to a vector representation. 
These representations from each component are then fed into an individual self-attention module for learning the similarities among the driving series. This scaled dot product self-attention scheme is defined as follows,
\begin{equation}\label{eq:DSANet/self_attention}
\begin{gathered}
    \mathbf{Z}^G = softmax(\frac{\mathbf{Q}^G(\mathbf{K}^G)^{\top}}{\sqrt{d_k}})\mathbf{V}^G,
\end{gathered}
\end{equation}
where $\mathbf{Q}^G, \mathbf{K}^G,$ and $\mathbf{V}^G$ are the set of queries, keys, and values obtained by applying projections to the output of global temporal convolution. $d_k$ represents the dimension of the keys. The position-wise feed-forward layer is defined as
\begin{equation}\label{eq:DSANet/position-wise}
\begin{gathered}
    \mathbf{F}^G = ReLU(\mathbf{Z}_O^{G}W_1 + b_1)W_2 + b_2,
\end{gathered}
\end{equation}
where $\mathbf{Z}_O^G$ is the final representation of the self-attention modules. In addition, it has a residual connection followed by layer normalization \cite{Ba2016layernorm} to make training easier and to improve generalization. To improve the robustness, an Autoregressive(AR) linear model is combined in a parallel manner because the scale of the output in non-linearity methods is not sensitive to the scale of the input.

Finally, the prediction of DSANet is obtained by combining self-attention and the AR prediction.

\subsection{L1 Trend Filtering}
\begin{figure}[!h]
\centering
\includegraphics[clip, trim={0.25cm 0 0 0}, width=0.5\textwidth]{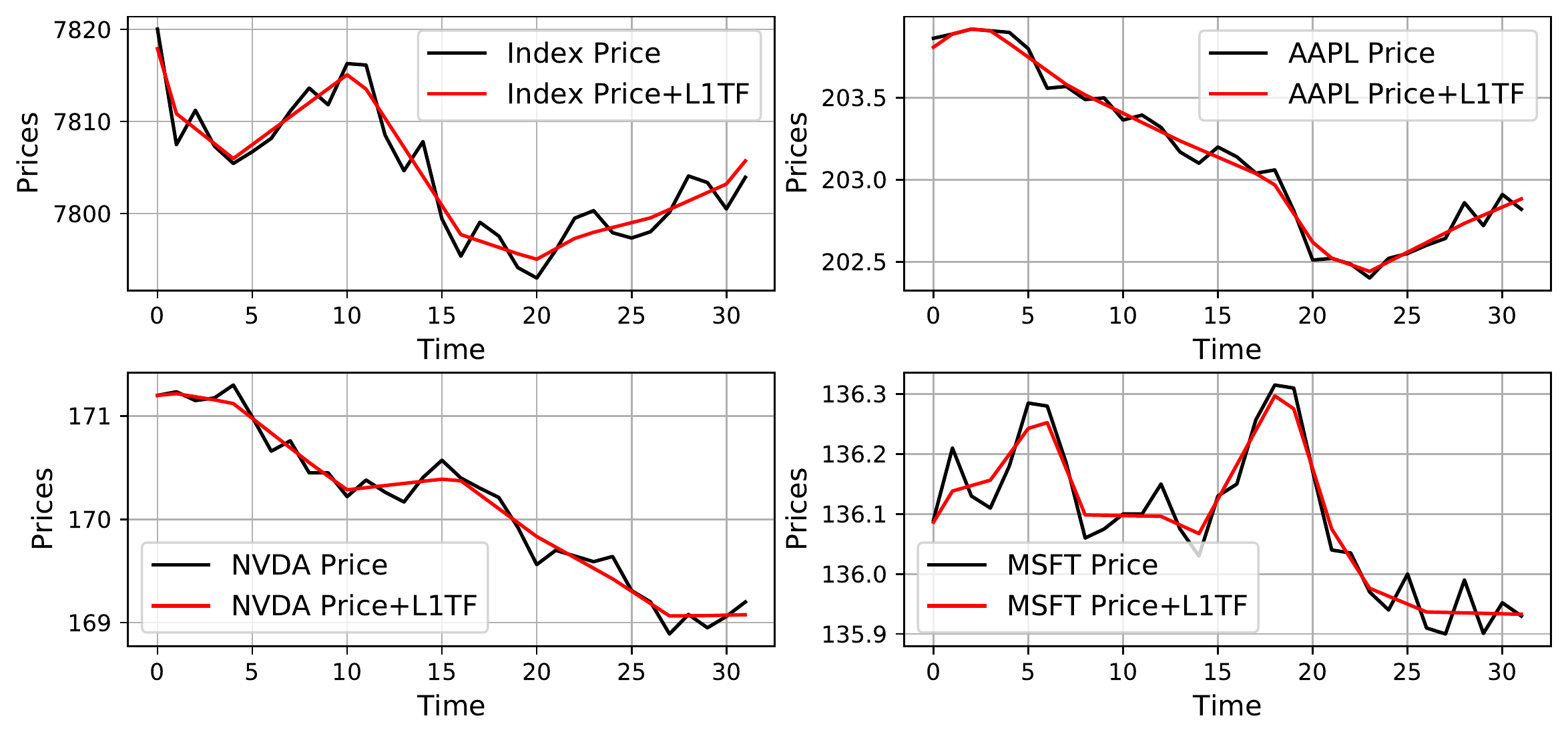}
\caption{Stock prices before and after the L1 Trend Filtering.}
\label{fig:l1tf_comparison}
\end{figure}

If a univariate time series $y_{t}$, $t=1,...,n$ it consists of trend $x_{t}$ that changes slowly and a random component $z_{t}$ which changes rapidly. The goal of L1 trend filtering is to estimate the trend component $x_{t}$ or, equivalently, estimate the random component $z_{t} = y_{t}-x_{t}$. This method can adjust $x_{t}$ to be smooth and can adjust $z_{t}$, referred to as the residual, to be small. The main principle is that the trend filtering produces trend estimates that are smooth in the sense that they are piecewise linear. This method allows the selection of the trend estimate as the minimizer of the objective function as follow, 

\begin{equation}\label{eq:l1tf}
\begin{gathered}
    (1/2)\sum_{t=1}^{n}{(y_{t}-x_{t})^2}+\lambda \sum_{t=2}^{n-1}\left | x_{t-1}-2x_{t}+x_{t+1} \right |,
\end{gathered}
\end{equation}
which can be written in matrix form as 
\begin{equation}\label{eq:l1tf_matrix}
\begin{gathered}
    (1/2)\left \| \mathbf{y}-\mathbf{x} \right \|_{2}^{2} + \lambda \left \| D\mathbf{x} \right \|_{1},
\end{gathered}
\end{equation}
where $ \left \| \mathbf{u}\right\|_{1} = \sum_{i} \left| u_{i} \right|$ denotes the $l_{1}$ norm of the vector $\mathbf{u}$ and $D\in\mathbb{R}^{(n-2)\times n}$ is second-order difference matrix. 
\begin{equation}\label{eq:dif_matrix}
\begin{gathered}
    D = 
    \begin{bmatrix}
    1 & -2 & 1 &   & &\\
      & 1 & 2 & -1 \\
      &   & \ddots & \ddots & \ddots \\
      &   &  & 1 & -2 & 1
    \end{bmatrix}
\end{gathered}
\end{equation}
In addition, $\lambda$ is a non-negative parameter used to control the trade-off between the smoothness of $x$ and the size of the residual. Fig. \ref{fig:l1tf_comparison} shows after the trend filtering that the fluctuating behavior of the original time series graph is stabilized and the trend filtering simply follows the general trend between two adjacent knot points. This means that time series models can detect some important signals more easily and that filtering simplifies the prediction of noisy time series. The important signal represents a change point (knot) when the time series trend changes. Many experiments in this paper show the effect of the trend filtering for three deep temporal neural network models.

\section{Experimental results}
We implement the proposed methods and the baseline models in the PyTorch framework. This sections describes five stock index datasets, after which we introduce the parameter settings of the  proposed methods and the evaluation metrics. Finally, we compare the proposed methods with different baseline methods, proving the effectiveness of our methods.

\begin{table}
    \begin{tabular}{cccc}
    \hline
    \multirow{2}{*}{Dataset}     & \multirow{2}{*}{Stock Exchange} & \multicolumn{2}{c}{Size} \\ \cline{3-4} 
                                 &                        & Train       & Test       \\ \hline
    NASDAQ 100 Stock             & NASDAQ                & 45,668       & 5,075       \\
    EURO STOXX 50 Stock          & Eurozone                & 65,349       & 7,261       \\
    Dow Jones Industrial Average & NYSE, NASDAQ                & 48,740       & 5,416       \\
    FTSE 100 Stock               & LSE                & 61,907       & 6,879       \\
    TSX 60 Stock                 & TSE                & 49,404       & 5,490       \\ \hline
    \end{tabular}
\captionsetup{justification=centering}
\caption{Information of the datasets.}
\label{tab:metr_la_table}
\end{table}

\subsection{Datasets}

To verify the effectiveness of the proposed methods, we utilize five stock market index datasets that represent the different stock exchanges from Bloomberg. These datasets are suitable for multivariate time series forecasting.
In order to reflect recent market trends, data from July of 2019 to January of 2020 are used here. Each dataset consists of a stock market index as the target of the prediction and the stock prices of companies as the input. We assume that the function used to compute the index out of the individual stocks is unknown. Thus, the deep neural network models must learn a predictive index function from observations. The frequency of the data is one minute. This satisfies the conditions of the deep neural networks, which require a large number of samples. Moreover, each dataset is divided into two cases: one with L1 trend filtering applied and the other with original inputs. The information of the datasets is shown in Table 1.

\begin{figure*}[!p]
\includegraphics[width=0.98\textwidth]{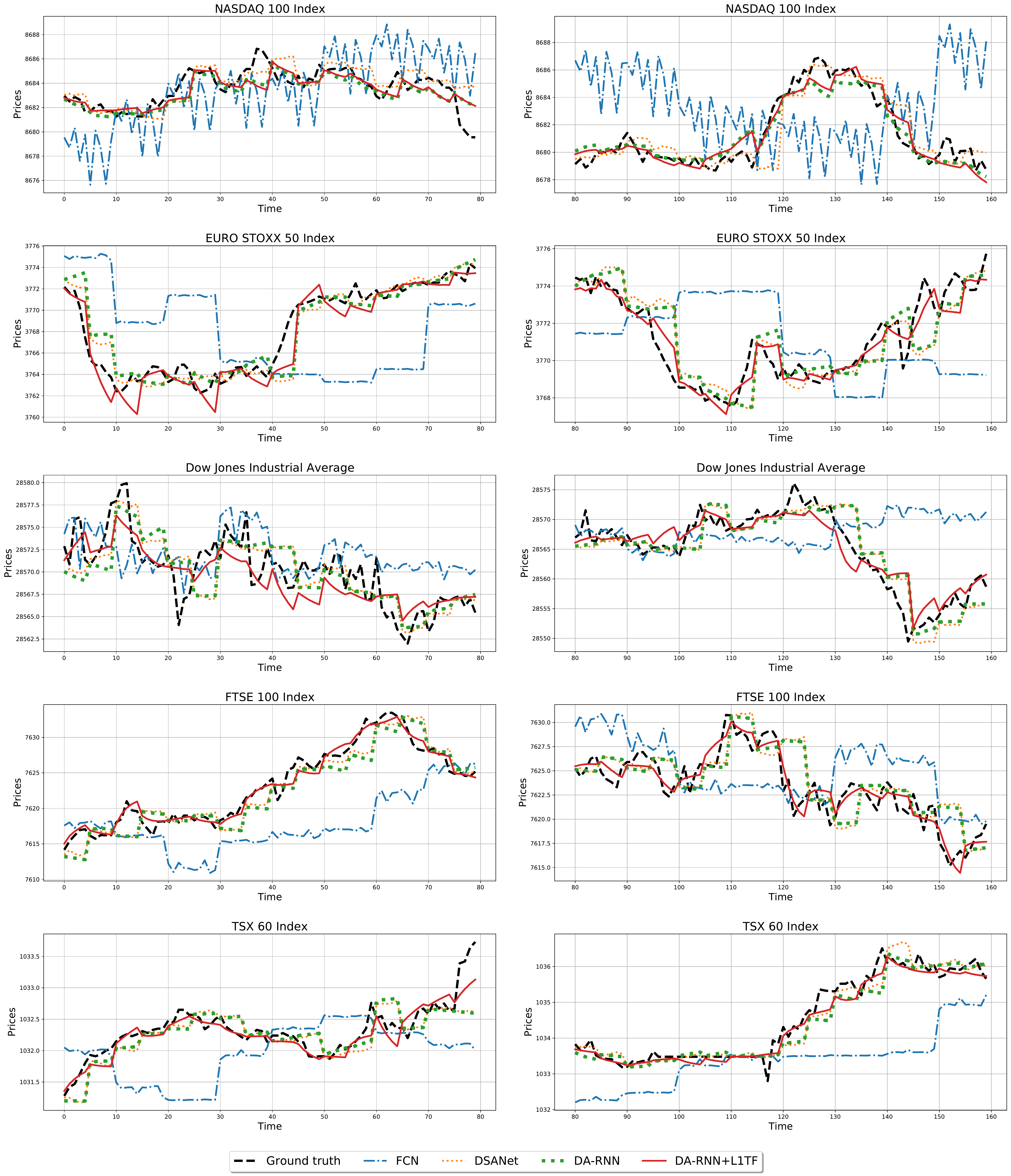}
\caption{Prediction results of DA-RNN+L1TF compared to other methods. Each dataset is located in each row having two different terms of time by representing two columns, and the prediction comparison from each figure is shown. As shown in the figure, the DA-RNN+L1TF outperforms other methods.}
\label{fig:whole_result}
\end{figure*}

\subsubsection{NASDAQ 100 Stock} The NASDAQ is a separate index of only 100 blue-chip companies listed on the NASDAQ index in the USA. This data covers the period of July 1, 2019 to January 10, 2020, $i.e.$, 193 days in total. In our experiments, we use the first 45,668 data points as the training set and the following 5,075 data points as the test set. The growth rate of the NASDAQ 100 Index during the period of test data is 3.26\%. 
\subsubsection{EURO STOXX 50 Index} The EURO STOXX 50 Index represents the 50 leading stocks in 12 Eurozone (\textit{e.g.}, Austria, France, Germany) countries and major sectors. It is calculated by the STOXX company. This data covers the period from July 1, 2019 to January 10, 2020, $i.e.$, 193 days in total. In our experiments, we use the first 65,349 data points as the training set and the following 7,261 data points as the test set. The growth rate of the EURO STOXX 50 Index during the period of test data is -3.47\%. 
\subsubsection{Dow Jones Industrial Average (DJIA)} The DJIA consists of only 30 blue-chip companies listed on the NYSE and NASDAQ indices in the USA. This data covers the period from July 2, 2019 to January 31, 2020, 195 days in total. In our experiments, we use the first 48,740 data points as the training set and the following 5,416 data points as the test set. The growth rate of the DJIA during the period of test data is 1.16\% .   
\subsubsection{Financial Times Stock Exchange (FTSE) 100 Index} The FTSE 100 index represents the stock prices of 100 companies in the order of market capitalization listed on the London Stock Exchange(LSE). The FTSE 100 Index is the leading index of the UK stock market. This data covers the period from July 1, 2019 to January 10, 2020, $i.e.$, 193 days in total. In our experiments, we use the first 61,907 data points as the training set and the following 6,879 data points as the test set. The growth rate of the FTSE 100 Index during the period of test data is -4.26\%.  
\subsubsection{Toronto Stock Exchange (TSX) 60 Index} The TSX 60 Index is a stock market index of 60 large companies in the order of market capitalization listed on the Toronto Stock Exchange (TSE). This data covers the period from July 1, 2019 to January 10, 2020, $i.e.$, 193 days in total. In our experiments, we use the first 49,404 data points as the training set and the following 5,490 data points as the test set. The growth rate of the TSX 60 Index during the period of test data is -0.16\%.  

\subsection{Parameters and Investment Model Settings}
To apply the L1 trend filtering feature to the input data, we set the parameter $\lambda$ to 0.005. In ARIMA, we set the parameters p, d, and q to corresponding values of 1, 0, and 0. There are four parameters in the DA-RNN: $i.e.,$ the number of input time steps in the window $T_i$, the number of output steps $T_o$, and the size of hidden states for the encoder m and decoder p. To find optimal parameters, we conducted a grid search. Finally, we find the best performance over the validation set which is used for evaluation when $T_i=64, T_o=5$, and $m = p\in\{64, 128\}$. For the parameter $T_o$, the higher the value of the parameter is, the more often a lagged prediction compared to the ground truth can be detected. Accordingly, this parameter value is fixed. For the other parameters $m$ and $p$, these parameters are modified proportionally relative to the number of driving time series for each dataset. In DSANet, the number of multi-head $n_{head}$ is set to 8 and both the inner-layer dimension of Position-wise Feed-Forward Networks and the output of the dimensions are modified proportionally to the number of driving time series for each dataset. In FCN, the parameter filters are 32 in all layers, and the kernel sizes are 7, 5, and 3 in the sequence layer. Since FCN does not perform well in our experiments, only the index value is used as an input. To compare the results between the model with the trend filtering feature and that without it, only the trend filtering feature is added with the index value.
To ensure that the prediction is properly done in the desired direction, a simple investment model that buys or sells only one amount for each trade is applied to the output of the regression model. The position is determined by comparing the last predicted values $\hat y_{T_o}$ with the actual value $y_{T}$.
Also, the initial balance is set proportionally to the initial price for each dataset. 

\subsection{Evaluation Metrics}
To measure the performance capabilities of various models for time series predictions, three different evaluation metrics are used: the root mean squared error (RMSE), the mean absolute error (MAE), and the mean absolute percentage error(MAPE), with the Adam optimizer. RMSE is defined as 
\begin{equation}\label{eq:RMSE}
\begin{gathered}
    RMSE = \sqrt{\frac{1}{N}\sum_{i=1}^{N}(y_{t}^{i}-\hat{y}_{t}^{i})^2},
\end{gathered}
\end{equation}
and MAE is defined as 
\begin{equation}\label{eq:MAE}
\begin{gathered}
    MAE = \frac{1}{N}\sum_{i=1}^{N}\left| y_{t}^{i}-\hat{y}_{t}^{i} \right|,
\end{gathered}
\end{equation}
where $y_{t}$ is the target at time $t$ and $\hat{y}_{t}$ is the predicted value at time t. MAPE is chosen as an evaluation metric because it can measure the proportion of prediction deviation in terms of true values. MAPE is defined as
\begin{equation}\label{eq:MAPE}
\begin{gathered}
    MAPE = \frac{1}{N}\sum_{i=1}^{N}\left| \frac{y_{t}^{i}-\hat{y}_{t}^{i}}{y_{t}^{i}} \right|.
\end{gathered}
\end{equation}
In the investment model, the rate of return, used as an evaluation measure, is computed by subtracting the initial balance from the previous balance, adding the value of the stocks an investor has, and dividing these numbers by the initial balance. 

\begin{equation}\label{eq:ReturnofRate}
\begin{gathered}
    Rate\ of\ Return(\%) = \frac{B_{Last}-B_{Initial} + Stocks\times P_{close}}{B_{Initial}}  \times 100,
\end{gathered}
\end{equation}
where $B_{Last}$ and $B_{Initial}$ are correspondingly the last balance and initial balance, and $P_{close}$ is the closing price of the index.

\begin{figure}[!t]
\includegraphics[clip, trim={0 0 0 0}, width=0.5\textwidth]{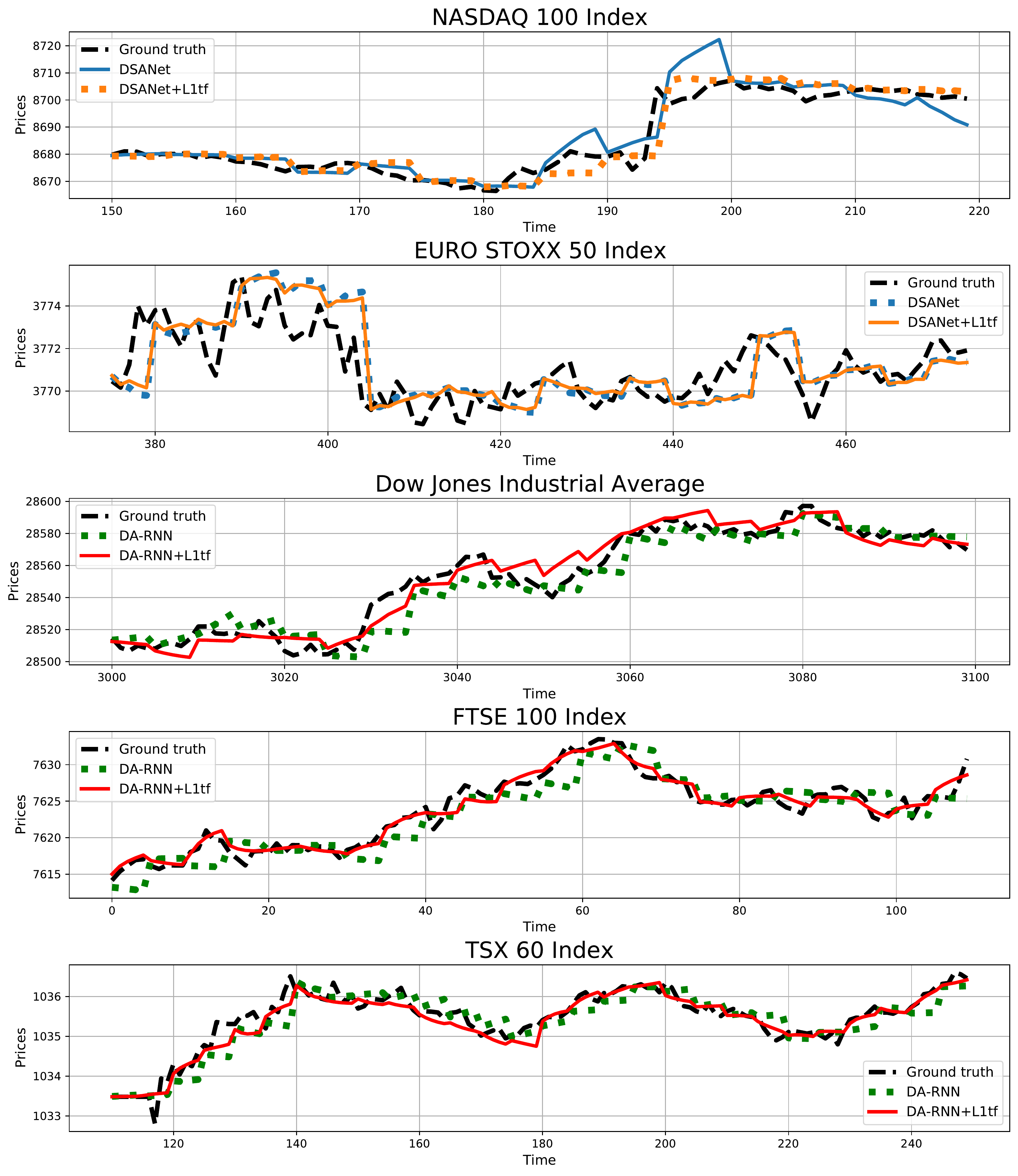}
\caption{Prediction results for the comparison between the model with the L1 trend filtering and without. Each dataset is located in each row, and each figure shows the prediction difference for two methods of the model.}
\label{fig:sample_result}
\end{figure}

\begin{table}[!t]
\scriptsize
\begin{tabular}{cccccc}
\hline
\multirow{2}{*}{\textbf{\begin{tabular}[c]{@{}c@{}}Dataset\\ (Growth Rate \%)\end{tabular}}} &
  \multirow{2}{*}{\textbf{Model}} &
  \multicolumn{4}{c}{\textbf{Metric}} \\ \cline{3-6} 
 &                        & \textbf{RMSE}    & \textbf{MAE}    & \textbf{MAPE}   & \textbf{Rate of Return (\%)} \\ \hline
\multirow{8}{*}{\textbf{\begin{tabular}[c]{@{}c@{}}NASDAQ \\ 100 Index\\ (3.26\%)\end{tabular}}} &
  \textbf{Lookahead} &
  252.65 &
  10.2805 &
  11.52 &
  0 \\
 & \textbf{ARIMA}         & 8.5656           & 4.4327          & 5.02287         & 1.01
                         \\
 & \textbf{FCN}           & 9.2603           & 5.0765          & 5.7540          & 7.02               \\
 & \textbf{FCN + L1TF}    & 8.8692           & 5.2586          & 5.9606          & 7.02               \\
 & \textbf{DSANet}        & 5.4383           & 2.6163          & 2.9638          & 6.99                        \\
 & \textbf{DSANet + L1TF} & 5.3484           & 2.5368          & 2.874           & \textbf{7.05}                        \\
 & \textbf{DARNN}         & 5.3079           & 2.6057          & 2.9514          & -3.93                         \\
 & \textbf{DARNN + L1TF}  & \textbf{5.2053}  & \textbf{2.5146} & \textbf{2.8486} & -0.71                         \\ \hline
\multirow{8}{*}{\textbf{\begin{tabular}[c]{@{}c@{}}EURO STOXX\\ 50 Index\\ (-3.47\%)\end{tabular}}} &
  \textbf{Lookahead} &
  85.681 &
  3.5038 &
  9.5152 &
  0 \\
 & \textbf{ARIMA}         & 3.8829           & 2.1286          & 5.6934          & 0.82                        \\
 & \textbf{FCN}           & 3.9643           & 2.2349          & 5.9765          & 1.08               \\
 & \textbf{FCN + L1TF}    & 3.4165           & 1.9012          & 5.0851          & 1.04               \\
 & \textbf{DSANet}        & 2.0377           & 1.1880          & 3.1777          & -1.85                        \\
 & \textbf{DSANet + L1TF} & 2.0049           & 1.1619          & 3.1077          & -0.18                        \\
 & \textbf{DARNN}         & 2.0232           & 1.1764          & 3.1469          & -0.14                        \\
 & \textbf{DARNN + L1TF}  & \textbf{1.8392}  & \textbf{0.9368} & \textbf{2.5057} & \textbf{7.39}                       \\ \hline
\multirow{8}{*}{\textbf{\begin{tabular}[c]{@{}c@{}}Dow Jones\\ Industrial Average\\ (1.16\%)\end{tabular}}} &
  \textbf{Lookahead} &
  787.98 &
  29.0741 &
  10.0784 &
  0 \\
 & \textbf{ARIMA}         & 19.7453          & 10.4543         & 3.6458          & -1.21                        \\
 & \textbf{FCN}           & 20.4051          & 10.8882         & 3.7968          & 9.83                        \\
 & \textbf{FCN + L1TF}    & 17.6532          & 9.7217          & 3.3899          & 9.83                        \\
 & \textbf{DSANet}        & 12.0051          & 6.3011          & 2.1974          & 9.69                        \\
 & \textbf{DSANet + L1TF} & 12.0281          & 6.3009          & 2.1973          & 12.41                        \\
 & \textbf{DARNN}         & 12.6665          & 6.6695          & 2.3258          & -0.75                        \\
 & \textbf{DARNN + L1TF}  & \textbf{11.5129} & \textbf{5.9758} & \textbf{2.0844} & \textbf{8.08}               \\ \hline
\multirow{8}{*}{\textbf{\begin{tabular}[c]{@{}c@{}}FTSE 100 Index\\ (-4.26\%)\end{tabular}}} &
  \textbf{Lookahead} &
  176.34 &
  7.1058 &
  9.6254 &
  0 \\
 & \textbf{ARIMA}         & 6.9347           & 3.8055          & 5.0598          & -0.55                         \\
 & \textbf{FCN}           & 6.9883           & 4.0133          & 5.3341          & -0.43                         \\
 & \textbf{FCN + L1TF}    & 6.2695           & 3.2788          & 4.3594          & -0.30                         \\
 & \textbf{DSANet}        & 3.7534           & 2.2669          & 3.0142          & -0.25                         \\
 & \textbf{DSANet + L1TF} & 3.7444           & 2.2648          & 3.0115          & 0.33                         \\
 & \textbf{DARNN}         & 4.244            & 2.3906          & 3.1788          & -0.24                         \\
 & \textbf{DARNN + L1TF}  & \textbf{2.9276}  & \textbf{1.3448} & \textbf{1.7874} & \textbf{5.75}               \\ \hline
\multirow{8}{*}{\textbf{\begin{tabular}[c]{@{}c@{}}TSX 60 Index\\ (0.16\%)\end{tabular}}} &
  \textbf{Lookahead} &
  27.976 &
  1.0155 &
  9.8054 &
  0 \\
 & \textbf{ARIMA}         & 0.6146           & 0.3508          & 3.3636          & 0.001                        \\
 & \textbf{FCN}           & 0.6678           & 0.3836          & 3.6775          & -0.003                        \\
 & \textbf{FCN + L1TF}    & 0.5886           & 0.3219          & 3.0861          & -0.003                        \\
 & \textbf{DSANet}        & 0.4606           & 0.2210          & 2.1185          & 0.65                        \\
 & \textbf{DSANet + L1TF} & 0.4584           & 0.2195          & 2.1044          & \textbf{0.67}                \\
 & \textbf{DARNN}         & 0.4673           & 0.2256          & 2.1627          & 0.09                        \\
 & \textbf{DARNN + L1TF}  & \textbf{0.4031}  & \textbf{0.1451} & \textbf{1.3917} & 0.3                        \\ \hline
\end{tabular}
\caption{Evaluation results of multivariate time series forecasting and Rate of Return. Compared with the original model, one applying L1 trend filtering has low values in terms of RMSE, MAE, and MAPE and is similar or higher in terms of Rate of Return.}
\label{tab:final_result}
\end{table}

\subsection{Lookahead model}
The Lookahead model is used for performance comparisons of both the prediction accuracy and Rate of Return. The prediction metric is calculated between the points of the last prediction and the current value. The rate of return is calculated identically to how it is done in a simple investment model. One difference between the Lookahead model and other models is that the predicted value $\hat{y}_{T_o}$ is not used for determining the current position. The Lookahead model only uses actual prices of test data to determine whether an investor buys or sells a stock.

\subsection{Prediction Results}
We compare the accuracy of Lookahead, ARIMA, FCN, FCN+L1TF (L1 trend filtering), DSANet, and DSANet+L1TF, DA-RNN, DA-RNN+L1TF models on all datasets. In addition, we conducted several experiments after creating two groups for each model, with one using L1 trend filtering, and the other using the original inputs.

Table \ref{tab:final_result} summarizes the evaluation results of all methods on the test set. In the "model" column in Table \ref{tab:final_result}, "model+L1TF" means that the model utilizes the trend filtering feature, as described above. For both the prediction and Rate of Return, we observe that the models with L1 trend filtering outperform the models without the trend filtering, indicating that the L1 trend filtering feature is helpful for proper predictions and that it achieves higher performances. In addition, considering that the models with the trend filtering feature obtain better results than the Lookahead model and the growth rate of each dataset, it can be shown that instances of noise in the time series are somewhat distinguishable from informative signals when we use models with the trend filtering feature. In Table \ref{tab:final_result}, DA-RNN+L1FT outperforms the other models in terms of the RMSE, MAE, and MAPE. In the FCN case, the larger the step sizes of the input and output are, the more unpredictable the output becomes. Although the DA-RNN and the DSANet show similar prediction results, DA-RNN+L1FT shows much better performance results compared to DSANet+L1TF. To show that DA-RNN+L1TF outperforms the other models, the prediction results for certain methods, including FCN, DSANet, DA-RNN, and DA-RNN+L1TF, are shown in Fig. \ref{fig:whole_result}. Only DA-RNN predicts well compared to the other methods 38, and 120 time steps on the FTSE 100 Index and at 48 and, 80 times steps on the TSX 60 Index.

Fig. \ref{fig:sample_result} shows comparisons of the prediction results between the models with the trend filtering feature and those without it. In the results with the NASDAQ 100 Index, after the price increases rapidly, DSANet+L1TF predicts the future index value insensitively compared to DSANet without the trend filtering feature. The trend filtering feature is also helpful to predict future time series appropriately in the DJIA, FTSE 100 index, and TSX 60 index results.

In addition, the paired t-test of two methods, with and without the trend filtering, shows the statistical significance of the fact that proposed method achieves better performances than models without filtering. On five datasets and three deep neural network models, twelve results of fifteen cases are statistically significant under significance level 0.05, and the results are shown in Table \ref{tab:t_test}.

\begin{table}[]
\begin{tabular}{ccccc}
\hline
Dataset                                                                                  & Model  & w/ L1TF & w/o L1TF & p-value \\ \hline
\multirow{3}{*}{\begin{tabular}[c]{@{}c@{}}NASDAQ\\ 100 Index\end{tabular}}              & FCN    & 8.8692  & 9.2603   & \textbf{0.007}   \\
                                                                                         & DSANet & 5.3484  & 5.4383   & 0.998   \\
                                                                                         & DARNN  & 5.2053  & 5.3079   & \textbf{0.027}   \\ \hline
\multirow{3}{*}{\begin{tabular}[c]{@{}c@{}}EURO STOXX \\ 50 Index\end{tabular}}          & FCN    & 3.4165  & 3.9643   & \textbf{0.000}   \\
                                                                                         & DSANet & 2.0049  & 2.0377   & \textbf{0.000}   \\
                                                                                         & DARNN  & 1.8392  & 2.0232   & \textbf{0.000}   \\ \hline
\multirow{3}{*}{\begin{tabular}[c]{@{}c@{}}Dow Jones \\ Industrial \\ Average\end{tabular}} & FCN    & 17.6532 & 20.4051  & \textbf{0.000}   \\
                                                                                         & DSANet & 12.0281 & 12.0051  & 0.466   \\
                                                                                         & DARNN  & 11.5129 & 12.6665  & \textbf{0.000}   \\ \hline
\multirow{3}{*}{FTSE 100 Index}                                                          & FCN    & 6.2695  & 6.9883   & \textbf{0.000}   \\
                                                                                         & DSANet & 3.7444  & 3.7534   & 0.380   \\
                                                                                         & DARNN  & 2.9276  & 4.244    & \textbf{0.000}   \\ \hline
\multirow{3}{*}{TSX 60 Index}                                                            & FCN    & 0.5886  & 0.6678   & \textbf{0.000}   \\
                                                                                         & DSANet & 0.4584  & 0.4606   & \textbf{0.013}   \\
                                                                                         & DARNN  & 0.4031  & 0.4673   & \textbf{0.000}   \\ \hline
\end{tabular}
\caption{Results of paired t-test between the models with L1TF and without L1TF. p-values with bold font indicate that the improvements of the models with L1TF over the vanilla models are statistically significant. Out of 15 cases (3 models and 5 datasets), 12 cases are statistically significant.}
\label{tab:t_test}
\end{table}

\section{Conclusion}
In this paper, we proposed a novel method that includes the L1 trend filtering feature which is helpful for the task of multivariate time series forecasting, especially for finance data with complex and non-linear dependencies. Furthermore, we applied this method to deep temporal neural networks that can detect certain important signals more easily and filtering simplifies the prediction of noisy time series to address the issue of the difficulty in distinguishing noise from informative signals. Experiments on five index datasets demonstrated that the proposed method outperforms deep neural networks that do not utilize our method for multivariate time series forecasting.

\begin{acks}
This work was supported by IITP grant funded by the Korea government (MSIT) (No.2017-0-01779, XAI and No.2019-0-00075, Artificial Intelligence Graduate School Program (KAIST))
\end{acks}

\bibliographystyle{ACM-Reference-Format}
\bibliography{sample_base}
\end{document}